\documentclass[runningheads]{article}

\usepackage{arxiv}

\usepackage[utf8]{inputenc} 
\usepackage[T1]{fontenc}    
\usepackage{hyperref}       
\usepackage{url}            
\usepackage{booktabs}       
\usepackage{amsfonts}       
\usepackage{nicefrac}       
\usepackage{microtype}      
\usepackage{lipsum}		
\usepackage{graphicx}
\usepackage{doi}

\usepackage{multirow}
\usepackage{amsmath}
\usepackage{amssymb}
\usepackage{latexsym}
\usepackage{placeins}

\renewcommand{\headeright}{Mehrtens et al., 2022}
\renewcommand{\undertitle}{Preprint}

\usepackage{graphicx}
\usepackage{tcolorbox}

\usepackage{wrapfig}

\usepackage[
backend=biber,
style=numeric,
sorting=none
]{biblatex}
\AtEveryBibitem{%
    {
     \clearfield{urlyear}}
    {}%
}
\begin{document}
\title{Pitfalls of Conformal Predictions for Medical Image Classification}
%
%
\author{ Hendrik Mehrtens, Tabea-Clara Bucher, Titus J. Brinker\\
	Division of Digital Biomarkers for Oncology \\ German Cancer Research Center (DKFZ) \\ Heidelberg, Germany \\
	\texttt{\{hendrikalexander.mehrtens\},\{tabea.bucher\}@dkfz-heidelberg.de} \\
	\texttt{titus.brinker@nct-heidelberg.de} \\
}

\maketitle              
\begin{abstract}

Reliable uncertainty estimation is one of the major challenges for medical classification tasks. While many approaches have been proposed, recently the statistical framework of conformal predictions has gained a lot of attention, due to its ability to provide provable calibration guarantees. 
Nonetheless, the application of conformal predictions in safety-critical areas such as medicine comes with pitfalls, limitations and assumptions that practitioners need to be aware of. We demonstrate through examples from dermatology and histopathology that conformal predictions are unreliable under distributional shifts in input and label variables. Additionally, conformal predictions should not be used for selecting predictions to improve accuracy and are not reliable for subsets of the data, such as individual classes or patient attributes. Moreover, in classification settings with a small number of classes, which are common in medical image classification tasks, conformal predictions have limited practical value.

\end{abstract}

\section{Introduction} \label{sec:Introduction}

In recent years, deep learning has gained popularity in the medical domain. Nonetheless, the adaptation into clinical practice remains low, as current machine learning approaches based on deep neural networks are not able to provide reliable uncertainty estimates \cite{begoli, laak, kompa2021}, which however are a necessary condition for the safe and reliable operation in and admission to clinical practice. 

Over the last years many publications in the machine learning literature have been focused on estimating the predictive uncertainty of machine learning models using a plethora of approaches, however, recent publications \cite{jaeger2023, mehrtens2022} have shown that these methods do not consistently outperform the baseline of the confidence value of a single neural network when evaluated over a range of tasks. Moreover, these approaches are heuristic and lack calibration \textit{guarantees}, which are essential in high-risk environments like medical classification.

Conformal Predictions (CP) \cite{vovk2005} have emerged as a promising framework for estimating uncertainty in neural network predictions, as they can provide a guaranteed level of true label coverage by constructing sets of predicted classes instead of single point predictions with uncertainty values, whereby larger sets indicate higher uncertainty. In this publication, we explore the viability and limitations of CP for medical image classification, where it has been used before, e.g. \cite{wieslander2021, olsson2022, lu2022, lu2022a, lu2022b}, addressing common misconceptions and pitfalls, underpinning our observations with experiments conducted in histopathological tissue and dermatological skin classification tasks. While some of these limitations are well-known in the CP literature, e.g. \cite{angelopoulos2022}\cite{lu2022}, the intend of this publication is to serve as a comprehensive guide for future practitioners that might aspire to use or learn about conformal predictions for their own work in medical image classification, to avoid misconceptions and pitfalls, that we encountered ourselves, in discussions with colleagues and found in applications in the literature. 

After briefly introducing the concept of CP, we focus on the guarantees offered by conformal predictions and the challenges they pose in medical scenarios, particularly in dealing with domain shifts. Another challenge is maintaining coverage guarantees under shifts in the label distribution, which are harder to control and observe compared to changes in the input domain. We discuss the distinction between marginal and conditional coverage and its consequences for the coverage of predictive set sizes, coverage of individual classes, and other data subgroups. In this context, we also explore the connection between conformal predictions and selective classification \cite{geifman2017} and why CP are insufficient for this task. Finally, we discuss the applicability and limitations of conformal predictions for medical image classifications, as here often tasks are considered that only have few classes or are even binary, for example, classification of benign and malignant tissue samples or the detection of a disease.

\section{Conformal Predictions} \label{sec:CP}

Conformal predictions (CP) \cite{vovk2005} is a low assumption posthoc calibration approach for probabilistic classification models. Given a classifier $\hat{Y}$, that was already optimized on a training dataset $D_{train}$, and a calibration dataset $D_{cal}$, CP forms a set of classes $C(\hat{Y}(x_i))$ out of the predictions of the model $\hat{Y}(x_i)$, so that

\begin{equation} \label{eq:marginal_coverage}
\mathbb{P}_{(x_i,y_i) \in D_{test}}(y_i \in C(\hat{Y}( x_i))) \geq  1- \alpha\;,
\end{equation} 
which is called the marginal coverage guarantee, where $1-\alpha$ is the desired coverage and the set-valued function $C$ is found on the calibration set.

Conformal predictions make no assumptions on the model or the data but only necessitate the property of exchangeability between the datasets $D_{cal}$ and $D_{test}$.
The set-valued function $C$ that forms predictive sets over the output of the classifier $\hat{Y}$ is found on the held out calibration data set $D_{cal}$ using a non-parametric quantile-based algorithm. Given the exchangeability of both data collections, the properties of this function then also hold for the test data set. 

Various approaches exist for forming the prediction sets using different conformity scores. In our experiments, we employ the adaptive prediction sets (APS) formulation \cite{romano2020}, previously used in image classification \cite{angelopoulos2022a}. While these approaches may employ different conformity scores to construct the predictive sets $C(\hat{Y}(x_i))$, they share the common goal of taking an initially uncalibrated uncertainty estimate from the trained classifier $\hat{Y}$ and finding a set of classes for each prediction that satisfies the expected marginal coverage guarantee based on the classifier's behavior on the calibration dataset and are by that nature susceptible to the in the following discussed pitfalls and limitations.

\section{Conformal Prediction for histopathology and dermatology} \label{sec:Application}

\begin{figure}
    
    \hspace{0.03\linewidth} \textbf{a)} CAMELYON17 \hspace{0.08\linewidth} \textbf{b)} HAM10K \hspace{0.16\linewidth} \textbf{c)} HAM10K\par
    
    \includegraphics[width=0.32\linewidth]{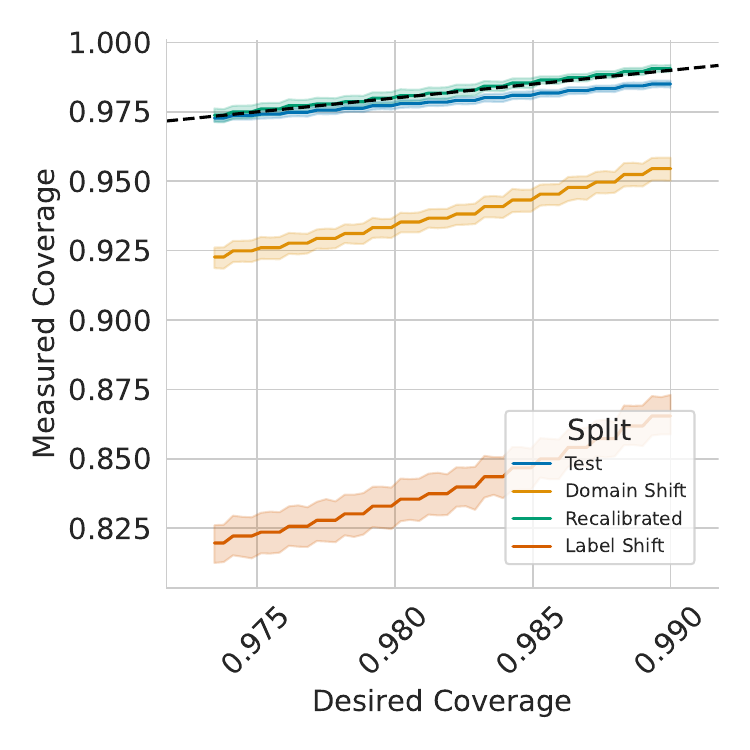}
    \includegraphics[width=0.32\linewidth]{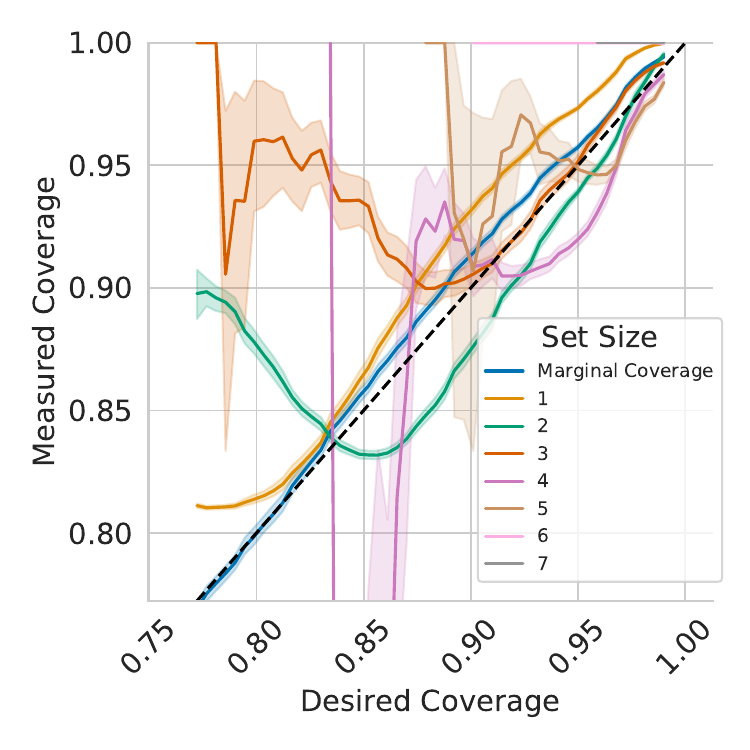}
    \includegraphics[width=0.32\linewidth]{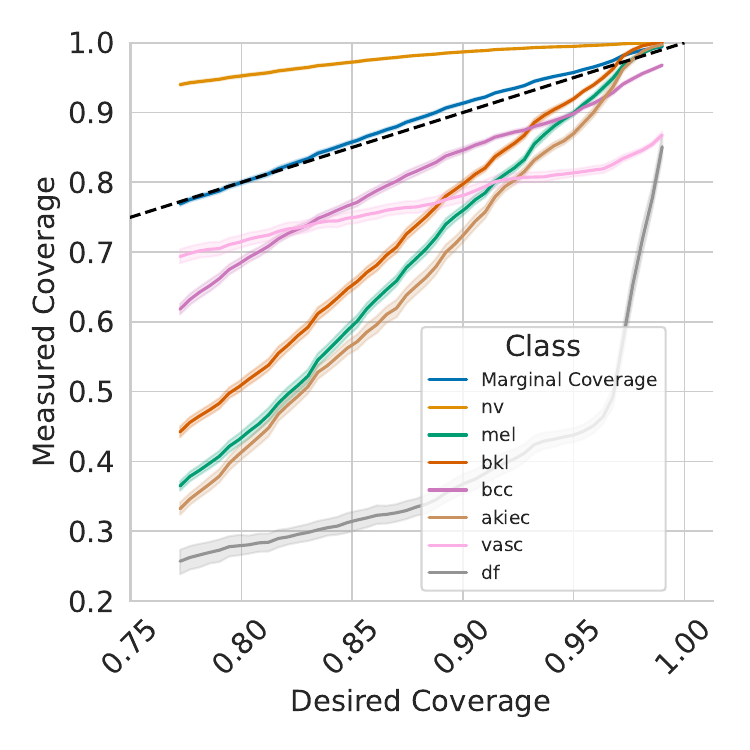}

    \caption{Coverage of the APS algorithm on two datasets under differing conditions. All curves are averaged over 5 runs and 10 sampled calibration sets. The dotted line shows the ideal calibration. a): Coverage on the CAMELYON17 dataset on the test set and under domain-shift and label distribution shift and after recalibrating on the domain-shifted data b): Coverage of different predictive set sizes on the HAM10K test dataset c): Coverage of different individual classes on the HAM10K test dataset.}
    \label{fig:coverage_experiments}
\end{figure}

\begingroup
\setlength{\intextsep}{0.15cm}%
\setlength{\columnsep}{0.6cm}%

\begin{wraptable}[11]{r}{7cm}
    \centering
    \begin{tabular}{l|c|c}

    Dataset & CAMELYON17 & HAM10K \\
    \hline
    Num. Classes & 2 & 7  \\
    Validation Set & 126672 & 967 \\
    Calibration Set & 1000 & 500 \\
    Accuracy (ID)& $97.56\%_{\pm 0.1}$ & $77.76\%_{\pm 0.4}$\\
    
\end{tabular}
    \caption{Overview of the used dataset, number of calibration points and the reached in-distribution accuracy over 5 evaluations}
    \label{tab:dataset_details}
\end{wraptable}

To showcase our observations we first trained five ResNet-34 \cite{he} on two medical image classification datasets each: HAM10K \cite{tschandl2018} for multi-class skin-lesion classification and \\ CAMELYON17 \cite{bandi2019} for histo-\\pathological whole-slide image (WSI) classification. The HAM10K dataset consists of seven skin conditions, while the CAMELYON17 dataset includes 50 WSIs with annotated tumor regions from five different clinics. We preprocessed the WSIs into non-overlapping patches of size 256x256 pixels. Our training setup followed the approach in \cite{mehrtens2022}, using the same augmentations and hyperparameters for both datasets. We trained our neural networks on Centers 1, 3, and 5 of the CAMELYON17 dataset and produced predictions for the Centers 2 and 4 for domain-shift experiments. It's important to note that conformal predictions have been previously employed in histopathology \cite{wieslander2021, olsson2022} and dermatology \cite{lu2022}. Our goal with these experiments is to showcase the applicability and limitations of conformal predictions for medical classification tasks, rather than achieving state-of-the-art classification results. We used singular neural networks for uncertainty estimation, as they have been shown to be sufficiently competitive \cite{jaeger2023, mehrtens2022} to more advanced approaches like Deep Ensembles \cite{lakshminarayanan2017} or Monte-Carlo Dropout \cite{gal, maddox}. \autoref{tab:dataset_details} provides details of the datasets and the achieved accuracies.

We utilized the APS algorithm \cite{romano2020, angelopoulos2022a} to compute the conformal prediction sets. We used 1000 calibration data points for CAMELYON17 and 500 for HAM10K, randomly sampled from their respective validation datasets, performing 10 calibration set samplings for each trained neural network, averaging the results over calibration sets and the five trained neural network per dataset.

The calibration curves for the APS algorithm are shown in \autoref{fig:coverage_experiments}, starting from the accuracy of the respective methods up to $100\%$ coverage. For the CAMELYON17 dataset we show the calibration on the test dataset, but also under domain-shift and label-shift. For the HAM10K dataset, we show the empirical coverage on the test set, but also the coverage of predictions of different conformal set sizes and different classes.

As depicted in the figures, conformal predictions reliably guarantee marginal coverage on the test sets for both the CAMELYON17 and HAM10K datasets. Further results will be discussed in subsequent chapters.
\endgroup

\section{Guarantees of conditional and marginal coverage} \label{sec:condmarginal}

The conformal predictions procedure ensures marginal coverage (\autoref{eq:marginal_coverage}), but not conditional coverage  $\mathbb{P}_{(x,y)\in D_{test}}(y \in C(\hat{Y}(x))| x) \geq 1-\alpha$. This means that while a conformal algorithm with a $1-\alpha=90\%$ marginal coverage guarantee provides a predictive set that covers the correct class with $90\%$ probability on average, it does not guarantee this for individual instances or structured subgroups of the data.

For example, in a dermatological classification task with a distribution of $95\%$ white patients and $5\%$ black patients, a $90\%$ coverage guarantee may provide a $92\%$ coverage to white patients, while systematically under-covering black patients with a $62\%$ coverage. Moreover, rare and potentially dangerous classes like melanoma can be under-covered, with frequent ones like skin nevi being over-covered, which is contrary to what should be desired in clinical practice.

\autoref{fig:coverage_experiments}.c shows the coverage of the different classes of the HAM10K dataset. The \textit{nevus} class is over-covered, all other classes, including the \textit{melanoma} class are undercovered, even though the overall coverage guarantee is met.

The literature proposes solutions to address these issues \cite{lu2022, vovk2012}, but they require separate calibration datasets for each combination of attributes (such as skin color, sex) and each class, which can be challenging for rare conditions or attribute combinations. 
Gathering such large calibration datasets may be difficult in certain settings, such as MRI or histopathological classification with rare conditions, and could limit the practical applicability of conformal predictions in such cases. Additionally, if such large datasets are available, there is a debate about whether they should be used to improve the classifier instead.

\begin{tcolorbox}[width=\textwidth]    
Marginal coverage is a property over the distribution and does not give guarantees for singular prediction sets.

Structured subgroups in heterogeneous datasets and imbalanced class-distributions might lead to false confidence in predictive sets.
\end{tcolorbox}    

\section{Conformal Predictions under Domain Shift} \label{sec:domain_shift}

The core assumption of conformal predictions is the exchangeability of calibration and test data sets. However, in practical deployments, this assumption might often be violated as validation sets are typically gathered together with the training data, while the deployment data can be encountered later in different contexts, for example in different clinics, with different image-acquisition processes. 

Another important pitfall for using conformal predictions, that can be easily overlooked, is to not only address shifts in the domain of the input variable, but also to consider changes in the label distribution, as the exchangeability assumption must hold for both. This pitfall can intuitively be missed by practitioners 
if the distribution of labels in the deployment domain differs from that of the training data, for example some diseases being more common in certain regions or populations.

We demonstrate both issues in \autoref{fig:coverage_experiments}.a using experiments conducted on the CAMELYON17 dataset. The coverage guarantees are successfully met when evaluating on the in-distribution test dataset. However, under domain-shift the coverage guarantee is not upheld. We generate an artificial test dataset by resampling the in-distribution dataset with a shifted label distribution. It is evident that the coverage guarantee is violated under these circumstances. This observation is supported by \autoref{fig:coverage_experiments}.c, which clearly shows that conformal predictions do not provide coverage guarantees for individual subsets of classes, making the coverage guarantee vulnerable to shifts in the class distribution.

Practitioners must ensure that the label distribution does not shift when the algorithm is deployed. If the distribution of labels in the deployment setting is known a-priori, for example through patient records of the past years, the calibration set should be resampled or weighted to match this distribution.

However, it is possible to gather a new representative calibration data set for each domain, which can then be used to recalibrate the model under domain shift.

This procedure is demonstrated in \autoref{fig:coverage_experiments}.a, where after re-calibrating on a subset of size 1000 of the Camelyon17 domain-shifted images, we again observe the coverage guarantees. 
As this has to be done for each application domain, this however might be a very large burden, especially combined with the fairness concerns discussed in \autoref{sec:condmarginal}.

\begin{tcolorbox}
    Conformal predictions guarantees do not hold under distributional-shift of the input variable \textbf{or} the label distribution. In this case, re-calibration on a new calibration dataset is required.
\end{tcolorbox}

\section{Conformal Predictions and Selective Classification} \label{sec:selectiveclassification}

\begin{figure}
    
    \hspace{0.03\linewidth} \textbf{a)} HAM10K 
    \hspace{0.16\linewidth} \textbf{b)} CAMELYON17 
    \hspace{0.08\linewidth} \textbf{c)} CAMELYON17\par

    \hspace{0.06\linewidth}  500 
    \hspace{0.27\linewidth} 1000 
    \hspace{0.25\linewidth} 10000\par

    \includegraphics[width=0.32\linewidth]{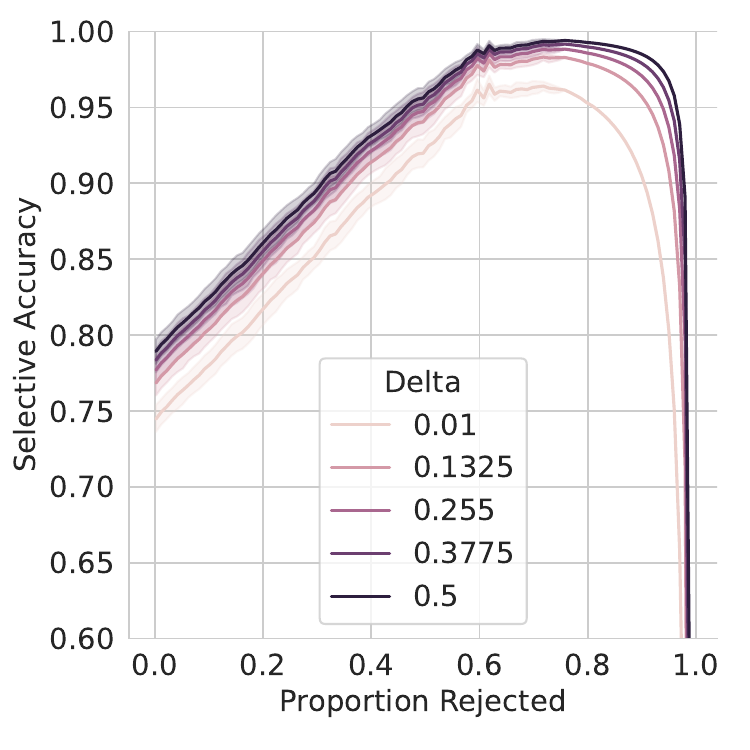}
    \includegraphics[width=0.32\linewidth]{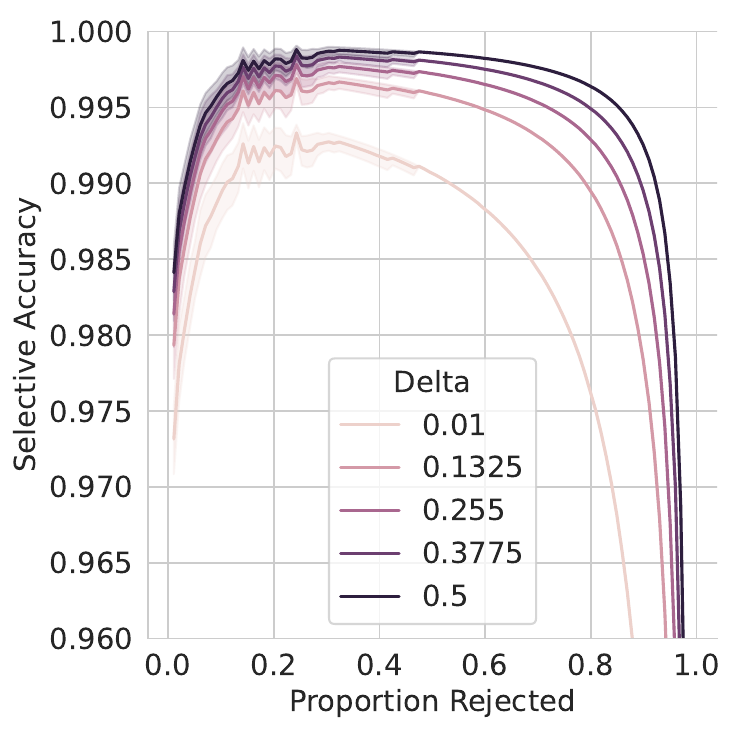}
    \includegraphics[width=0.32\linewidth]{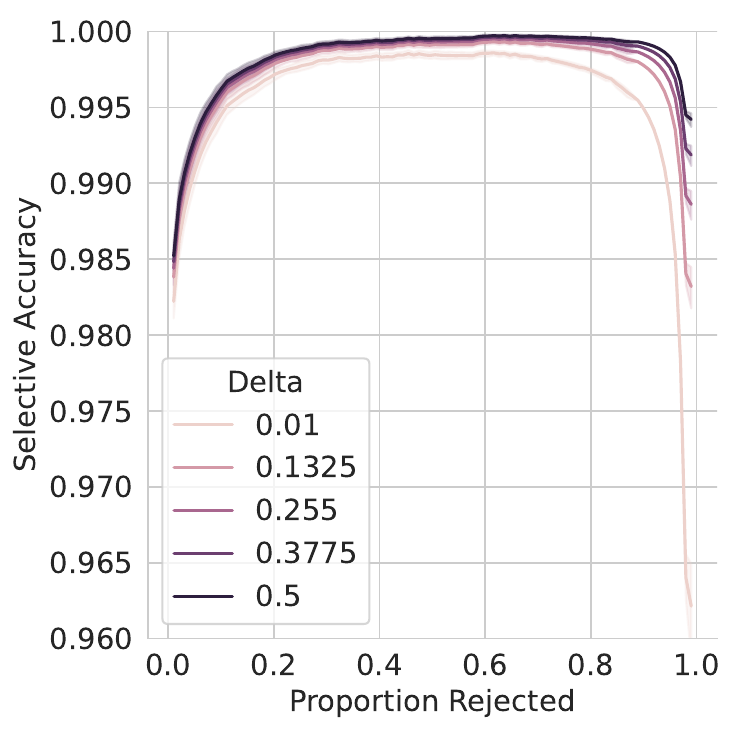}

    \caption{Selective Accuracy curves, using the approach of \cite{angelopoulos2022}, on HAM10K with 500 calibration points (left) and CAMELYON17 with 1000  points (middle) and 10000 (right) calibration points. We plot the curves for different values of $\delta$, the failure rate of the algorithm, ranging from 0.01 to 0.5. All curves are averaged over 5 runs.}
    \label{fig:selective_classification}
\end{figure}

We have come across several publications \cite{lu2022a, olsson2022} that utilize conformal predictions (CP) to enhance the classification accuracy of a classifier by solely using predictive sets of size 1. However, we argue that this is an incorrect application of CP, as it offers no control over the accuracy of predictions in sets of size 1 and cannot guarantee that these predictions will have higher accuracy than the rest of the predictions. While in practice, predictive sets of size 1, which represent the most \textit{certain} predictions of the neural network, often achieve higher accuracy than the rest of the predictions, this is a general property of the underlying uncertainty measurement of the neural network, not the conformal predictions procedure.

\autoref{fig:coverage_experiments}.b illustrates the measured coverage of different predictive set sizes. As can be seen, while the overall coverage is nearly ideally calibrated, the coverage of the individual set sizes can vary significantly, as CP does not guarantee any level of coverage over subsets of the data. As previously expected, the set size 1 predictions are in fact over-covered. However, CP does not offer control over the coverage of this subset or its size and therefore offers no additional value, compared to directly just utilizing the most confident predictions directly.

In recent years multiple publications have shown that neural networks are well capable of selective classification \cite{jaeger2023, geifman2017, mehrtens2022} and as such this framework should be employed when selecting predictions for higher accuracy. With this approach, one can obtain high-probability guarantees for desired accuracy levels, while conformal predictions cannot provide any coverage guarantees for prediction sets of size 1. Angelopoulos et al. \cite{angelopoulos2022} provide a statistical framework for selective classification in their worked examples. Their procedure guarantees the following property with a chosen probability $1-\delta$, where $\lambda$ is dependent on $\delta$:

 \begin{equation}
     \mathbb{P}_{(x,y)\in D_{test}}(y = argmax(\hat{Y}(x))| max(\hat{Y}(x)) \geq \lambda) \geq 1-\alpha
 \end{equation}
 
Contrary to CP, the approach produces point predictions instead of prediction sets and the number of required data points is dependent on the model, dataset, and the distribution of produced uncertainty estimates.

\autoref{fig:selective_classification} shows this approach for the HAM10K and CAMELYON17 datasets, for different failure probabilities ($\delta$) of the algorithm.
The reachable accuracy level and the required number of calibration data points are dependent on the task and the trained model. The lower the tolerated failure rate ($\delta$), the more conservative the estimate of the reached accuracy. After a certain percentage of rejected data points, the guaranteed accuracy of each curve drops, as the distribution of estimated accuracy values becomes too wide, due to too few remaining data points. However, with more calibration data points, higher accuracies are reachable, as can be seen for the CAMELYON17 curve on the right with 10000 calibration images. The CAMELYON17 predictions require far more calibration datapoints to see any increase in estimated accuracy after a $40\%$ rejection rate, which demonstrates the task dependency.

\begin{tcolorbox}
    Conformal predictions can neither guarantee nor control accuracies on the prediction sets of small cardinality. If selective classification is desired, practitioners can use more appropriate frameworks.
\end{tcolorbox}

\section{Classification with few classes} \label{sec:binary}

\begingroup
\setlength{\intextsep}{0.15cm}%
\setlength{\columnsep}{0.15cm}%

\begin{wrapfigure}{!RT}{0.5\textwidth}
    \includegraphics[width=0.49\linewidth]{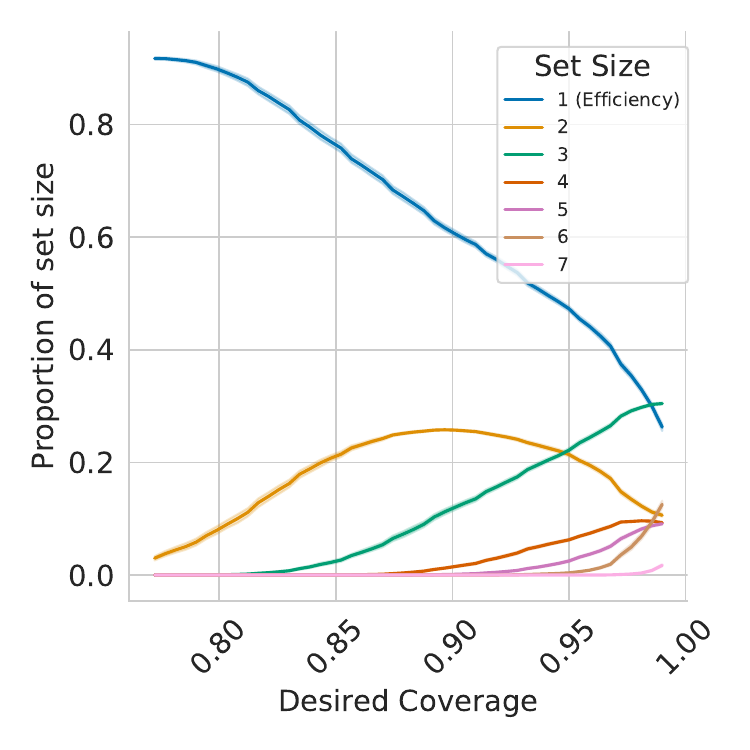}
    \includegraphics[width=0.49\linewidth]{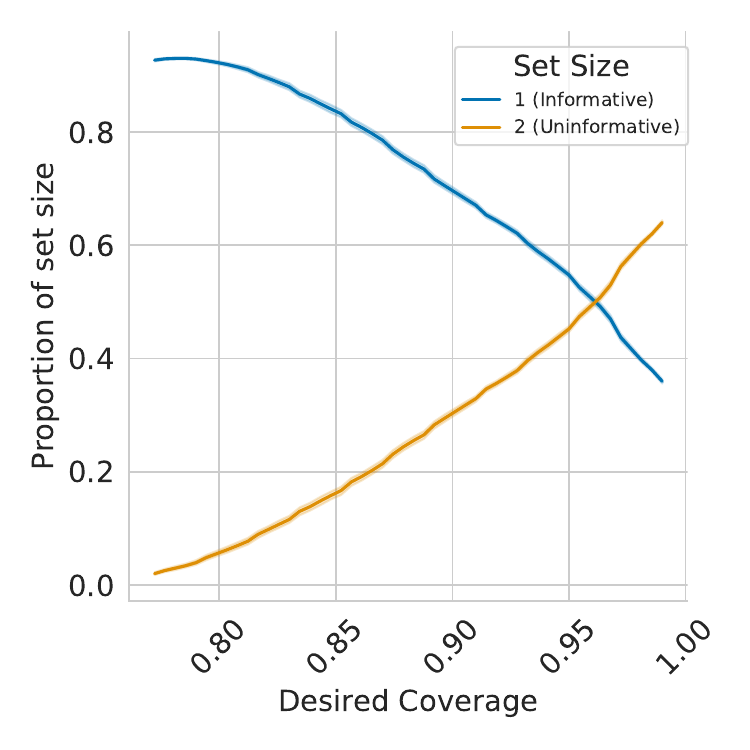}
    \caption{Efficiency of HAM10K with the seven individual classes (left) and classes mapped to benign and malignant (right).}
    \label{fig:efficiency}
\end{wrapfigure}

Medical classification tasks often only have a limited number of classes, contrary to some natural image classification tasks, for example, the ImageNet classification challenge \cite{deng2009}.
Many of these tasks are binary classification tasks, e.g. detecting the presence of tumors or other illnesses, and in others, the classes can be categorized into relevant super-classes, for example, benign and malignant conditions.
A binary classification setting is an extreme case for CP, due to the low number of possible uncertainty quantifications as there are only 3 possible prediction sets, both classes individually and the combination of both. A prediction set encompassing \textit{both} classes, or sets of classes containing benign and malignant classes, can be uninformative to the clinician. Therefore, an algorithm that produces many of these prediction sets, while faithfully ensuring the desired coverage, lacks practical value. With a rising desired level of coverage, the efficiency \cite{angelopoulos2022} of the CP algorithm drops, leaving the practitioner only with the most certain predictions, of set size 1, the same situation discussed in \autoref{sec:selectiveclassification}. 

As an example consider a binary classification case with a desired marginal coverage of 90\%. This could be met with (80\% correct single predictions, 10\% both, 10\% false single predictions) or (90\% both, 10\% false single predictions).
Conformal predictions do not provide control over the outcome, and the practical value of each case can vary significantly.
\autoref{fig:efficiency} that with rising guarantees, the proportion of set size 1 predictions declines. Even though there are seven classes, when the classes are mapped to benign and malignant conditions, which are the most relevant for treatment decisions, most of the predictions are uninformative.

\endgroup
\begin{tcolorbox}
    The practical value of conformal predictions in settings with few classes or categories can be low, due to the coarse resolution of prediction sets. 
\end{tcolorbox}

\section{Conclusion} \label{sec:conclusion}

Conformal predictions provide a valuable statistical framework for obtaining guarantees on uncertainty estimates, addressing the otherwise heuristic nature of neural nework uncertainty estimates. However, especially in situations encountered in the medical classification context, they have limitations, assumptions, and pitfalls that practitioners should be aware of. 
In this publication, we have presented an overview of these limitations, pitfalls, and interpretations of conformal predictions guarantees in the context of often situation in medical image classification. Solutions to challenges like fairness, imbalance in class coverage, and domain shifts in input and label space exist but require additional calibration data, which may not always be available in sufficient quantities. Moreover, consideration should be given to whether this additional data would be better utilized for improving the classifier. Additionally, in medical classification tasks with few classes, conformal predictions may have limited practical value. Conformal predictions should not be utilized to perform selective classification, as other, better suited frameworks exist.
 We hope that this publication will help practitioners navigate potential pitfalls and address common misunderstandings associated with conformal predictions.

\newpage

\section{Acknowledgements}
This publication is funded by the 'Ministerium für Soziales, Gesundheit und Integration', Baden Württemberg, Germany, as part of the 'KI-Translations-Initiative'.
 Titus Josef Brinker owns a company that develops mobile apps (Smart Health Heidelberg GmbH, Heidelberg, Germany), outside of the scope of the submitted work.
\printbibliography

\end{document}